\documentclass[10pt, a4paper]{article}

\usepackage[final]{lrec2026} 
\usepackage{expex}
\usepackage{xcolor}
\definecolor{transcolor}{gray}{0.45}
\newcommand{\trans}[1]{\textcolor{transcolor}{\textit{#1}}}
\title{Coconstructions in spoken data:\\ UD annotation guidelines and first results}

\name{
\begin{tabular}{c}
Ludovica Pannitto$^1$, Sylvain Kahane$^2$, Kaja Dobrovoljc$^3$, \\
Enela Battaglia$^1$, Bruno Guillaume$^4$, Caterina Mauri$^1$, Eleonora Zucchini$^5$
\end{tabular}
}

\address{
$^1$~University of Bologna, Bologna - Italy;\\
$^2$~Université Paris Nanterre, Modyco, Paris - France; \\
$^3$~University of Ljubljana and Jozef Stefan Institute, Ljubljana - Slovenia;\\
$^4$~Université de Lorraine, CNRS, Inria, LORIA, Nancy - France; \\
$^5$~Masaryk University, Brno - Czech Republic\\
}

\abstract{The paper proposes annotation guidelines for syntactic dependencies that span across speaker turns — including collaborative coconstructions proper, wh-question answers, and backchannels — in spoken language treebanks within the Universal Dependencies framework.
Two representations are proposed: a speaker-based representation following the segmentation into speech turns, and a dependency-based representation with dependencies across speech turns. New propositions are also put forward to distinguish between reformulations and repairs, and to promote elements in unfinished phrases.\\
\newline
\Keywords{spoken language, coconstruction, backchannel, reformulation, syntactic treebank, French, Italian, Slovenian} }

\begin{document}

\maketitleabstract

\section{Introduction}

Spoken language is intrinsically temporal, incremental, and interactive (e.g.,~\citealt{hopper1992times,auer_online-syntax_2024,deppermann2015temporality}): unlike written text, it unfolds in real time and is jointly produced by multiple participants cooperating in the construction of syntactic structure. Linearized transcriptions, however, necessarily flatten this temporal and multi-party dimension.

Universal Dependencies (UD, \citealt{de_marneffe_universal_2021}), despite its success as a cross-linguistic annotation framework, inherits a broadly shared assumption in syntactic theory: that syntax operates within the boundaries of a sentence. While this assumption is largely unproblematic for written language (although sentence boundaries detection is not yet considered a completely solved task, e.g.~\citealt{rudrapal_sentence_nodate,read_sentence_2012, redaelli_is_nodate}), it raises difficulties for spoken interaction, where segmentation into sentence-like units is neither trivial nor theoretically neutral.

Two additional issues complicate the picture. First, segmenting the speech flow is not equivalent to segmenting speaker turns: different segmentation strategies may be adopted depending on whether the focus is on individual speakers’ production or on jointly constructed syntactic units. Second, syntax can be viewed both as a cognitive mechanism for building grammatical structures and as an interactional resource for coordinating meaning between speakers. These perspectives stem from different research traditions, yet an annotation framework such as UD should not commit exclusively to either one.

We introduce a novel proposal for annotation guidelines that can enable resource creators and users to account for syntactic dependencies holding among speech produced by different speakers.
In Section~\ref{sec:spokentreebanks}, we argue that spoken data require an explicit treatment of syntactic relations that cross turn boundaries. We refer to these relations as \emph{coconstructions}.  We provide a typology of such constructions in Section~\ref{sec:typology}, and we propose annotation guidelines that preserve information from both speaker-based and dependency-based perspectives, while remaining fully compatible with the UD formalism, in Section~\ref{sec:guidelines}.
Section~\ref{sec:conversion} presents the conversion between the speaker-based and the dependency-based representation. A quick overview of the implementation of our annotation scheme in three treebanks from French, Italian, and Slovenian is provided in Section~\ref{sec:treebanks}.

\section{The units of spoken UD treebanks}
\label{sec:spokentreebanks}

No clear agreement emerges in the literature about what a \textit{sentence} (as in, maximal unit of segmentation) in spoken language is~\citep{sacks1974simplest, pietrandrea_notion_2014, mettouchi2020prosodic} and, despite sentences being the main building blocks of UD resources, no explicit definition of what should count as a sentence is ever given in the current annotation guidelines.

Syntactic annotation frameworks typically operate on segmentation units that resemble sentences but can vary greatly in their definitions. This mismatch becomes particularly evident when syntactic dependencies span across speaker turns, overlap with other speakers’ contributions, or are resumed after interruptions.
\citet{pietrandrea_notion_2014} recall three commonly applied criteria employed for the definition of sentences, namely the locutionary criterion, that attributes sentences to the locutionary responsibility of a given speaker; the graphic or prosodic criterion, that defines sentences based on punctuation markers or major prosodic breaks; and the syntactic criterion, that focuses on sentences as maximal structurally autonomous units, linked to the surrounding context by discursive relations.

While for standard written data the three criteria are likely to converge to the same analysis, this is not the case for spoken language and other varieties such as user-generated content~\citep{sanguinetti_treebanking_2023}.
Earlier attempts at building spoken treebanks annotated for constituent syntax have implemented a turn-based approach. For instance, in the Penn Treebank-style annotation of the Switchboard corpus, turns by different speakers are parsed as standalone trees~\citep{bies1995bracketing}; in the extended Switchboard Dialog Act Corpus - SwDA~\citep{jurafsky1997switchboard}, by representing a discourse layer on top of a syntactic one, only some forms of co-construction, for instance collaborative completions, receive a tag.
Another example is the Verbmobil corpus, where the primary domain of syntactic analysis and annotation is ``the dialog turn, (…) an uninterrupted contribution by one dialog participant''~\citep[108]{hinrichs2000verbmobil}; while this approach can account for some phenomena of spoken language such as disfluencies (see~\citealt{meteer1995disfluency}), it privileges  single-speaker syntax and obscures collaborative syntax, which cannot be properly searched within trees.
Dependency frameworks are, as a matter of fact, more apt to represent relations holding across turn boundaries. However, among existing spoken language treebanks in UD, no uniform segmentation criterion is applied, probably as a result of pre-existing design choices in the source data~\cite{dobrovoljc_spoken_2022}.

We argue that two different units of analysis need to be considered in the case of spoken data:
\begin{itemize}
    \item \textbf{speaker-dependent maximal units}, loosely related to the notion of \textit{turn} and that anchor segmentation to the speech production of each speaker participating in the conversation, as it is usually done;
    \item \textbf{structurally-autonomous maximal units}, based on the notion of government~\citep{tesniere1959elements, tesniere2015elements, blanche1987pronom}, defined by~\citet{kahane_typologie_2012} as the property by which words impose constraints on other words, specifically constraints on (i) their syntactic category, (ii) their morphological or syntactic marking, and (iii) their linear position.
\end{itemize}

An alternative segmentation strategy, adopted for instance in Rhapsodie, is based on the notion of illocutionary unit, defined through prosodic and pragmatic criteria. Such units capture the organization of speech in terms of illocutionary completeness and prosodic packaging, and therefore reflect an important dimension of spoken interaction. However, because they are grounded in prosodic-pragmatic segmentation, they do not necessarily align with domains of syntactic dependency. As a result, using illocutionary units as primary units for syntactic annotation would make the representation of syntax dependent on a specific analysis of prosodic and illocutionary organization, which is not always available, consistently applicable, or comparable across corpora and languages. By contrast, speaker-dependent maximal units provide a more neutral and operational point of departure: they are directly anchored in the observable distribution of speech across participants, preserve information about production and turn organization, and can be systematically enriched to recover syntactic dependencies that extend across unit boundaries, without conflating speaker organization, prosodic packaging, and syntactic structure into a single segmentation criterion.

In the remainder of this paper, we refer to speaker-dependent maximal units as \textit{maximal units} and to structurally-autonomous maximal units as \textit{rectional units}.
Currently, spoken treebanks in UD are organized in speaker-dependent units, and only Rhapsodie (see Section~\ref{sec:rhapsodie}) and KIParla Forest (see Section~\ref{sec:kiparla}) provide special features for cases of coconstructions.

\section{A typology of coconstructions}
\label{sec:typology}

We consider as \emph{coconstructions} those syntactic relations (without restrictions on dependency type) that hold between utterances produced (i) by different speakers or (ii) by the same speaker after an interruption by another participant. In interactive scenarios, speakers may jointly contribute to the realization of a single syntactic structure, sometimes in a tightly coordinated fashion and sometimes with partial or even complete overlap~\citep{lerner_syntax_1991, 10.1007/978-3-662-03293-0_3,Sacks1992, helasvuo2004shared, Calabria_2026}. From a dependency-based perspective, this implies that syntactic relations cannot always be confined to a single speaker turn.

Our notion of coconstruction is grounded in a dependency-oriented view of syntax and is compatible with the concept of rectional unit (\emph{unité rectionnelle}) introduced by~\citet{benzitoun2010tu,kahane_typologie_2012}, described in Section~\ref{sec:spokentreebanks}: this is defined independently of speaker turns or prosodic segmentation and it captures syntactic governance relations wherever they occur in the speech flow. This makes the notion particularly well suited to spoken interaction, where the realization of a single rectional domain may be distributed across multiple speakers or interrupted by other contributions. Coconstructions can thus be understood as cases in which a single rectional unit is jointly realized across turns, by one or more speakers.

On this basis, we distinguish three major types of coconstructions, reflecting different interactional and syntactic configurations: \emph{co-construction proper}, \emph{wh-question constructions}, and \emph{backchannelling}.

\subsection{Coconstruction proper}
\label{sec:coconstruction-proper}

Coconstruction proper covers cases in which multiple contributions target the same dependency structure by completing, duplicating, or extending dependencies. Such contributions can be produced (i) by different speakers or (ii) by the same speaker after an interruption by another speaker, i.e. in non-contiguous utterances, which formally results in the same pattern as (i)\footnote{We do not treat the case of contiguous maximal units produced by the same speaker, as we hypothesize that these should be merged in a single unit in case they can be linked through syntactic dependency, thus yielding a single maximal rectional unit.}.
In the remainder of the article we will refer to Speaker~1 and Speaker~2 as the producers of the first and second unit which are focus of the discussion. Here and after, in the event of an interruption, Speaker~2 coincides with Speaker~1. All examples are reproduced preserving the original transcription conventions of the corpus.

The key distinction within coconstructions proper concerns whether Speaker~2 realizes a dependency that is already projected by Speaker~1 (case A), or whether Speaker~2 adds a new dependency to the structure initiated by Speaker~1 (case B).

\paragraph{A. Speaker 2 realizes an already projected dependency.}
When Speaker~2 realizes a dependency projected by Speaker~1, we distinguish three subcases reflecting the status of that dependency at the moment of realization: unrealized, in progress, or already realized.

\begin{description}
    \item[A.1 Completion:] Speaker~1 leaves a syntactic dependency unrealized, and Speaker~2 provides the missing material, \textit{completing} the structure initiated by Speaker~1, as in Example~(\ref{ex:stable});
    \ex
    \label{ex:stable}English (\citealt[p. 72]{10.1007/978-3-662-03293-0_3})\\
    L: his position is \textbf{pretty uh}\\[2pt]
    A: … \textbf{stable}.
    \xe
    \item[A.2 Parallel realization:] The same syntactic dependency is realized at the same time by both speakers. Speaker~1 projects and eventually completes the structure, often after hesitation or a false start, while Speaker~2 simultaneously or near-simultaneously fills the same syntactic slot, as in Example~(\ref{ex:stiamoinsieme}), see also Figure~\ref{fig:stiamo insieme} in Appendix~\ref{sec:appendix}. This configuration frequently results in overlap;
    \ex
    \label{ex:stiamoinsieme}Italian (KIParla\footnote{\cite{Mauri_2019}. In all examples from KIParla, square brackets indicate overlaps between speakers' productions.}, BOD2018)\\
    BO140: mh: vabbè [ci \textbf{stiamo] insie:me:}\\
    \phantom{BO140:} \trans{well we stay together there}\\
    \phantom{BO140:} esatto: c'è il divano comodo,\\
    \phantom{BO140:} \trans{right there's a comfortable sofa}\\[2pt]
    BO118: \textbf{[state insie:me:]} \\
    \phantom{BO118:} \trans{stay together}
    \xe
    \item[A.3 Subsequent realization:] Speaker~1 completes their own syntactic structure, and Speaker~2 subsequently selects a dependency in Speaker~1’s structure and realizes it again, either by repeating or replacing material in the same syntactic slot (as in Examples~(\ref{ex:bourgeoise}) and (\ref{ex:finzione})).
    \ex
    \label{ex:bourgeoise}French (Rhapsodie, D2001)\footnote{Curly brackets indicate conjuncts in a stacking construction. See Section~\ref{sec:stacking}. The sign // indicates the end of an illocutionary unit. The sign //+ indicated that there is a link to the next unit.}\\
    \$L2 ``eh bien" je crois que je ne me suis pas \\
    \phantom{\$L2} \trans{well I think I haven't}\\
    \phantom{\$L2} conduit d'une façon conforme à ce qu'on \\
    \phantom{\$L2} \trans{behaved as one would}\\
    \phantom{\$L2} attend ``euh" \{ \{ \textbf{d'une jeune fille d'abord} \\
    \phantom{\$L2} \trans{expect from a young girl first}\\
    \phantom{\$L2} \textbf{ | \^{} et d'une femme ensuite} \} |\} //+ \\
    \phantom{\$L2} \trans{and from a woman afterwards}\\[2pt]
    \$L1 \{| \textbf{d'une jeune bourgeoise} |\} ?//+\\
    \phantom{\$L1} \trans{from a young bourgeois girl?}\\[2pt]
    \$L2 \{| \textbf{``disons" d'une jeune bourgeoise }\} //\\
    \phantom{\$L2} \trans{let's say from a young bourgeois girl}
    \xe
    \ex
    \label{ex:finzione}Italian (KIParla, KPS021)\\
    PKP126: {quindi l- la linea \textbf{tra finzione e realtà}}\\
    \phantom{PKP126:} \trans{so the line between fiction and reality}\\
    \phantom{PKP126:} cioè tra verità non verità \\
    \phantom{PKP126:} \trans{that is between true and false}\\
    \phantom{PKP126:} non ho ancora capito [dove sta]\\
    \phantom{PKP126:} \trans{I still haven’t understood where it lies}\\[2pt]
    PKP125: {\textbf{[più tra de]tto e non detto}}\\
    \phantom{PKP126:} \trans{more between said and unsaid}
    \xe
\end{description}

\paragraph{B. Speaker 2 adds a new dependency.}
Speaker~2 realizes a syntactic dependency that is \textit{not} already projected by Speaker~1, thereby \textit{extending} the syntactic structure initiated by Speaker~1 (Examples~(\ref{ex:know}), (\ref{ex:parentesi}), (\ref{ex:yes}))

\ex
\label{ex:know}English (\citealt[p. 81]{10.1007/978-3-662-03293-0_3})\\
M: they must know each other.\\[2pt]
H: … \textbf{very well}.
\xe
\ex Italian (KIPArla, BOA3017)\\
\label{ex:parentesi}
BO139: quando si parlano sopra \\
\phantom{BO139:} \trans{when they speak one over the other}\\
\phantom{BO139:} devi mettere delle parentesi quadre \\
\phantom{BO139:} \trans{you have to put square parentheses}\\
\phantom{BO139:} intorno \textbf{alle parole} \\
\phantom{BO139:} \trans{around words}\\[2pt]
BO146: ah okay \\
\phantom{BO146:} \trans{oh okay}\\[2pt]
BO139: \textbf{che si sovrappongono} \\
\phantom{BO139:} \trans{that overlap}\\[2pt]
BO136: \textbf{come nei sottotitoli}\\
\phantom{BO136:} \trans{like in subtitles}
\xe

\subsection{\textit{Wh-}questions}

Answers to \textit{wh}-questions are analysed as syntactically dependent on the question, and thus as part of the same rectional unit, as already proposed in the guidelines of Rhapsodie.

\ex
\label{ex:climb}Italian (KIParla, PTA007)\\
TOR001: \textbf{dove} vai ad arrampicare? \\
\phantom{TOR001} \trans{where do you climb?}\\[2pt]
TOI007: \textbf{al bi side\footnote{`bi side' is the quasi-phonetic rendering of the name of a climbing gym, `Bside'.}} vicino alla colletta \\
\phantom{TOI007:} \trans{at the bi side close to colletta}
\xe

We therefore consider answers to \textit{wh}-questions as cases of coconstructions, insofar as they realize syntactic material licensed by the interrogative structure of the preceding turn (see an example in Figure~\ref{fig:boule}). By contrast, we do not treat answers to polar questions as coconstructions, since their response items do not depend syntactically on any element of the question nor fill a pre-existing syntactic slot (ex., cf. Example~(\ref{ex:climb}) with A: \textit{do you climb at the bi side?} B: \textit{yes}).

\subsection{Backchannelling}
\label{sec:backchanneling}

Backchannels are short productions uttered by one participant while another speaker holds the floor. They may be verbal or paraverbal (e.g.\ \emph{mhm} as in Example~(\ref{ex:mhmh}), \emph{yes} as in Example~(\ref{ex:yes}), \emph{right}) and mainly serve to signal attention or alignment~\citep{ward2000prosodic,mereu2024backchannels, furko2020discourse, khan2023role}. At their core, these verbal reactions show that the speaker has heard the contribution of the partner, often adding that it has been understood and accepted.

To be recognized as backchannels, such utterances (\textit{i}) must be addressed to the content of the other speaker’s contribution;
(\textit{ii}) must not be required or expected by the preceding turn (e.g. answers to \textit{wh-} and polar questions are expected and required, so they cannot be considered backchannels, see~(\ref{ex:climb}));
(\textit{iii}) must not require a reaction from the main speaker.

\ex
\label{ex:mhmh}Italian (KIParla, BOA3017)\\
BO145: ma perché mamma c'ha dei \\
\phantom{BO145:} \trans{but because my mom has}\\
\phantom{BO145:} pregiudizi nei miei confronti \\
\phantom{BO145:} \trans{prejudices against me}\\[2pt]
BO139: \textbf{mhmh} \\
\phantom{BO139:} \trans{mhmh}\\[2pt]
BO145: e poi daniela non devi avere \\
\phantom{BO145:} \trans{and then daniela you shouldn't have}\\
\phantom{BO145:} pregiudizi su di me \\
\phantom{BO145:} \trans{prejudice against me}
\xe

\ex
\label{ex:yes}Italian, (KIParla, BOD2018)\\
BO118: sì sì ma anch'io però era proprio \\
\phantom{BO118:} \trans{yeah yeah same here but I really just}\\
\phantom{BO118:} l'esigenza di stare da sola\\
\phantom{BO118:} \trans{needed to be alone}\\[2pt]
BO140: \textbf{sì }di stare da sola di non \\
\phantom{BO140:} \trans{yeah to be alone not}\\
\phantom{BO140:} parlare \\
\phantom{BO140:} \trans{to talk}
\xe

Backchannels differ from co-construction proper and from \textit{wh}-questions coconstructions, in that they do not contribute syntactic material to the rectional unit projected by the main speaker. Nevertheless, they are tightly integrated into the temporal unfolding of speech and must be accounted for in the syntactic analysis of spoken interaction. \citet{jucker1996explicit} argue that the distinction between a ``discourse marker'' and a ``backchannel'' is often more about conversational role than linguistic function and both belong to a broader category of elements used to negotiate ``common ground''.

We are aware that the syntactic status of backchannels is not uncontroversial. What counts as ``syntax'', however, is itself a matter of theoretical tradition: different frameworks draw the boundary between syntax, pragmatics, and interaction differently, and UD itself already encodes relations — such as \texttt{discourse} — that would not be considered syntactic under a narrow structural definition. More importantly, even if a backchannel is not itself a syntactic dependent, it actively shapes the syntactic structure being built by the main speaker: it may prompt continuation, signal that repair or elaboration is needed, or mark the boundaries within which the syntactic unit unfolds. Ignoring backchannels in the annotation of spoken interaction thus risks misrepresenting the conditions under which syntactic structure is produced and interpreted. Including them in the \texttt{MISC} field — without altering any dependency relation — is a minimal and non-disruptive way of making this interactional layer visible to downstream users of the treebank.

\section{Proposed guidelines}
\label{sec:guidelines}

To explicitly represent coconstructed configurations in UD annotation, we introduce two features encoded in the \texttt{MISC} field: \texttt{Coconstruct} and \texttt{Backchannel}. These features do not replace dependency relations, but provide a lightweight mechanism to signal linkage between speaker-based maximal units involved in coconstruction phenomena.

All features introduced in this paper are encoded in the \texttt{MISC} field, keeping the proposal fully backward-compatible: the core dependency structure, POS tags, morphological features, and tokenization remain untouched. The \texttt{MISC} field is already used across treebanks for typographic information (\texttt{SpaceAfter=No}), transliteration, named entity spans, and discourse-level features; our proposal follows this established practice.

As a general principle, both \texttt{Coconstruct} and \texttt{Backchannel} are always assigned to the element that comes second in time: the feature is assigned to the responding token and its value identifies the eliciting or projecting token in the earlier utterance. This also enables multiple later tokens to point to the same earlier anchor.

Formally, \texttt{Coconstruct} encodes a backward pointer of the form \texttt{$\langle$deprel$\rangle$::$\langle$sent\_id$\rangle$::$\langle$tok\_id$\rangle$}, while \texttt{Backchannel}, which already uniquely identifies a dependency relation (i.e., \texttt{discourse:backchannel}), encodes a backward pointer of the form \texttt{$\langle$sent\_id$\rangle$::$\langle$tok\_id$\rangle$} (see Figure~\ref{fig:apostrofo}).

The \texttt{Coconstruct} feature is used in two main configurations, which differ in the syntactic status of the dependency slot targeted by the responding token:
\begin{description}
    \item[Completion] of an open dependency slot: the responding token realizes a dependency that was projected but not realized in the preceding unit.
    \item[Stacking] on an already realized dependency slot: the responding token provides an alternative realization of a dependency that is already present in the preceding unit.
\end{description}

\subsection{Completion of an open dependency slot}

A configuration qualifies as completion of an open dependency if all of the following conditions are met:
\begin{itemize}
    \item Speaker~1 produces a token that syntactically licenses a dependent (although not necessarily requiring it), but the dependent is not realized before Speaker~2’s intervention.
    \item Speaker~1 does not complete the same dependency slot before or concurrently with Speaker~2’s contribution.
\end{itemize}

When these requirements are met, the head of the coconstructed dependency is the token in Speaker~1’s utterance that projects the unrealized dependency. Head identification follows standard UD principles: the head is the lexical item that selects the dependent.

This configuration applies both to cases A.1 and B described in Section~\ref{sec:typology}.
In cases of A.1 Speaker~1 might produce an element that projects a dependency but remains syntactically unfinished, because its canonical dependent is not realized before turn transfer. This situation is particularly frequent with adpositions (more specifically, prepositions), complementizers, and similar selecting elements. From a UD perspective, such cases raise a problem: the projected dependent is missing, but dependency annotation cannot be postponed or left underspecified.
To address this, we adopt the promotion mechanism used in UD for ellipsis. When a projected dependency remains unrealized in Speaker~1’s utterance, the unfinished element is promoted to fill the syntactic position that would normally be occupied by its missing complement.
This allows the dependency structure to remain well-formed while preserving information about the element’s incomplete status.

We introduce two additional \texttt{MISC} features to encode this mechanism: \texttt{Scrap=Yes} to mark any element (or subtree) whose syntactic projection remains unfinished and \texttt{Promotion=$\langle$deprel$\rangle$} to indicate the dependency relation the element would have had before promotion, i.e. the relation it would bear to its governor if it were present.

\ex\label{ex:mestiere}%
Italian, (KIParla, BOA3017)\\
BO146: ((ride)) vabbè te ti ci ti ci devo \textbf{fare} \\
\phantom{BO146:} \trans{((laughs)) well you I have to do you}\\
\phantom{BO146:} \textbf{pe::r} \\
\phantom{BO146:} \trans{by}\\[2pt]
BO139: ((ride)) \textbf{mestiere} \\
\phantom{BO139:} \trans{((laughs)) trade}
\xe

For instance, in Example~(\ref{ex:mestiere}), a noun is missing after the preposition \textit{per}.
The oblique prepositional phrase is unfinished, so the \texttt{obl} dependency relation has to go on the preposition.
The preposition, which should have been the \texttt{case} of the missing noun, has then been promoted. Since this promotion follows a \texttt{case} relation, the preposition receives the feature \texttt{Promotion=case}, as well as a feature \texttt{Scrap=Yes} indicating that the phrase it heads is unfinished (see first line in Figure~\ref{fig:mestiere}).

Unfinished syntactic structures may arise in spoken interaction independently of whether coconstruction is eventually realized.
The use of \texttt{Scrap=Yes} therefore does not presuppose that the missing material will be supplied later: it encodes structural incompleteness as such, regardless of subsequent interactional developments.
\texttt{Promotion} is essentially used in cooccurrence with \texttt{Scrap=Yes} when there is an unfinished phrase without its expected head.\footnote{\texttt{Promotion} could, for instance, be introduced in case of gapping, as in (i) where \textit{department} is promoted as root of the second conjunct.

\vspace{3pt}
\noindent (i) \ Mary called the police and Peter the fire department.

\vspace{3pt}
The feature \texttt{Promotion=obj} could be an elegant way to indicate that \textit{department} is \texttt{obj} of an ellipsed verb.
}
As we show in Section~\ref{sec:conversion}, the feature \texttt{Promotion} plays an important role in the conversion to the dependency-based representation.

The promotion of an element to a given position introduces an additional problem, because a dependent of a promoted word can be either a true dependent of the word itself or of its missing head. We mark this by means of a feature \texttt{Head}, with possible values \texttt{Word} and \texttt{Position}, to solve this problem. In Example~(\ref{ex:very}), \textit{nice} will be promoted as the \texttt{obj} of \textit{has} and it receives a feature \texttt{Promotion=amod}, \textit{a} is \texttt{det} of \textit{nice} with a feature \texttt{Head=Position}, and \textit{very} is \texttt{advmod} of \textit{nice} with a feature \texttt{Head=Word}. In appendix, the full annotation of Example~(\ref{ex:very}) is given (Figure~\ref{fig:nice}) together with a Slovenian example (Figure~\ref{fig:sl}) taken from the corpus.

\ex\label{ex:very}%
A: she has a very nice \\[2pt]
B: attitude
\xe

\subsection{Stacking on an already realized dependency slot}
\label{sec:stacking}

A configuration qualifies as stacking on an already realized dependency slot if all of the following conditions are met:
\begin{itemize}
   \item Speaker~1 realizes a complete dependency structure, including the dependent that fills the relevant syntactic slot;
   \item Speaker~2 produces material that is syntactically comparable to the dependent already realized by Speaker~1 and could occupy the same dependency position.
\end{itemize}

In stacking configurations,
Speaker~2 does not complete an open dependency, but rather produces material that targets the same syntactic slot as an existing dependent. From a production perspective, Speaker~2’s contribution is not structurally autonomous: its interpretation requires reference to Speaker~1’s syntactic structure.

Stacking can however be instantiated by various deprels.

The most straightforward case is that of simple repetitions or repairs without any communicative intention, for which we recommend the \texttt{reparandum} relation.
These are cases such as \textit{des, des} or \textit{ce dont, ce dont} in Example~(\ref{ex:French}), or cases where a word is unfinished as \textit{il sotto\textasciitilde} ‘the sub\textasciitilde’ in Example~(\ref{ex:sottotitolo}) (see also Figure~\ref{fig:repair}): the first speaker has difficulties to produce the word \textit{sottotitolatore} ‘subtitler’, which is then correctly produced by the second speaker.
\texttt{reparandum} is a right-headed relation in UD: in order to mark a coconstruction on the latter item of the relation, we introduce the special deprel \texttt{repair} which is converted to \texttt{reparandum} in dependency-view.

\ex\label{ex:sottotitolo}%
Italian (KIParla, BOA3017)\\
BO147: infatti te potresti fare \textbf{il sotto\textasciitilde} \\
\phantom{BO147:} \trans{indeed, you could be the subti\textasciitilde}\\
\phantom{BO147:} quello che fa i sottotitoli \\
\phantom{BO147:} \trans{the one that does subtitles}\\[2pt]
BO145: \textbf{sottotitolatore} \\
\phantom{BO145:} \trans{subtitler}
\xe

If instead the first conjunct is complete and the second conjunct is different, we consider that we have a reformulation.
Example~(\ref{ex:finzione}) is a typical case: The first speaker already reformulates \textit{tra finzione e realtà} ‘between fiction and reality’ by \textit{tra verità non verità} ‘between true and false’ and then the second speaker proposes a second reformulation \textit{più tra detto e non detto} ‘more between said and unsaid’.

We propose to introduce a new relation for reformulation, which we call \texttt{conj:reform}. Contrary to coordinated conjuncts, where conjuncts have different referents, in a reformulation, the conjuncts are different denotations of the same referent~\citep{kahane_typologie_2012}.
We extend the relation \texttt{conj:reform} to other cases where two conjuncts target the same referent.

The first case is confirmation. In Example~(\ref{ex:bourgeoise}),
speaker \$L2 produces \textit{d'une jeune fille d'abord et d'une femme ensuite} ‘from a young girl first and from a woman afterwards’ and \$L1 proposes the reformulation \textit{d'une jeune bourgeoise} ‘from a young bourgeois girl’ (with an interrogative prosody). Then \$L2 confirms this proposition by repeating it (with the addition of the discourse marker \textit{disons} ‘let's say’).
The second case is the answer to a wh-question as in Example~(\ref{ex:climb}), where the interrogative pronoun \textit{dove} ‘where’ is instantiated by the answer \textit{al bi side} ‘at the bi side’.
A third case is correction. We will illustrate it by the monologue in Example~(\ref{ex:French}). In this Example, the speaker repeats three times \textit{des Français} ‘of the French’, so there is no reformulation strictly speaking, but the second occurrence questions the choice of the proposed denotation, while the third, preceded by \textit{pas seulement} ‘not only’, refutes it.

In all cases, \texttt{$\langle$tok\_id$\rangle$} points to the head of the first denotation.

\ex French (Rhapsodie, D0004)\label{ex:French}\\
\$L1: mais ce dont, ce dont vous parlez, c'est\\
\phantom{\$L1:} \trans{but what, what you're talking about is}\\
\phantom{\$L1:} la crise générale des, des Français, oui,\\
\phantom{\$L1:} \trans{the general crisis of the French, yes,}\\
\phantom{\$L1:} enfin, des Français, pas simplement \\
\phantom{\$L1:} \trans{well, of the French, not just}\\
\phantom{\$L1:} des Français, hein, des, de l'humanité\\
\phantom{\$L1:} \trans{of the French, but of the, of humanity}\\
\phantom{\$L1:} et de la lecture.\\
\phantom{\$L1:} \trans{and reading.}
\xe

\section{Automatic conversion between speaker-based and dependency-based}
\label{sec:conversion}

The two proposed annotation views (speaker-based and dependency-based) rely on different maximal segmentation principles.
As UD and SUD annotation schemas strongly rely on the notion of `sentence' for segmentation, we propose providing the two views separately.
The speaker-based view, properly enriched with the features introduced in Section~\ref{sec:guidelines}, can be converted into the dependency-based view, which relies on the definition of `sentence' as `rectional unit', enabling coconstructions to be realised as regular syntactic dependencies.

\begin{figure*}
  \centering
  \includegraphics[scale=0.45]{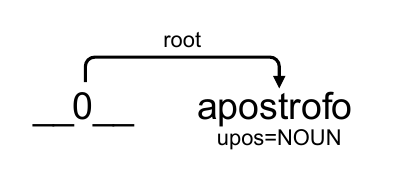}
  \includegraphics[scale=0.45]{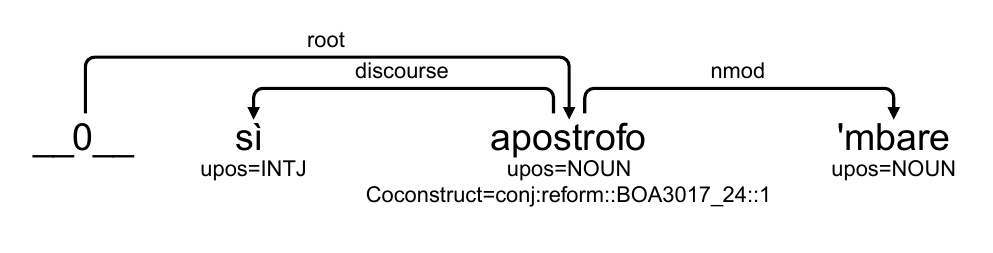}
  \includegraphics[scale=0.45]{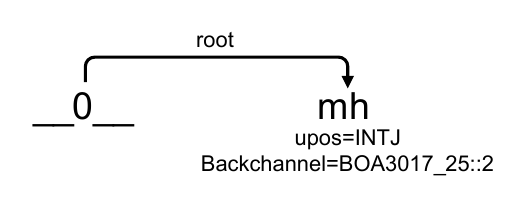}\\
  \includegraphics[scale=0.45]{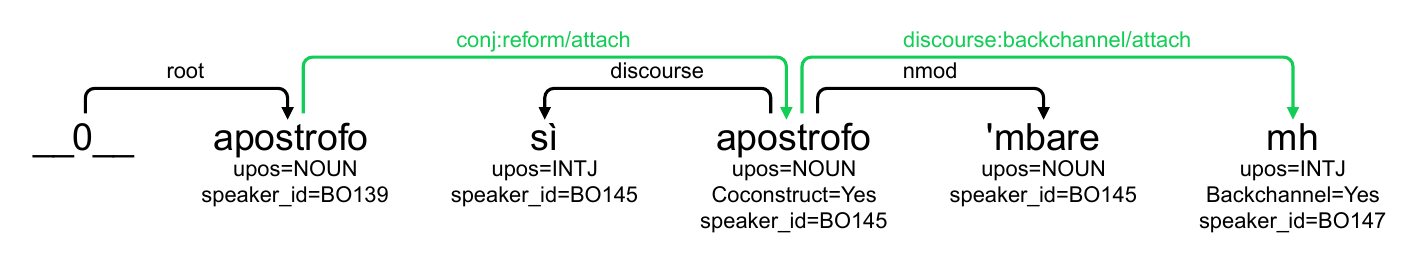}
  \includegraphics[scale=0.45]{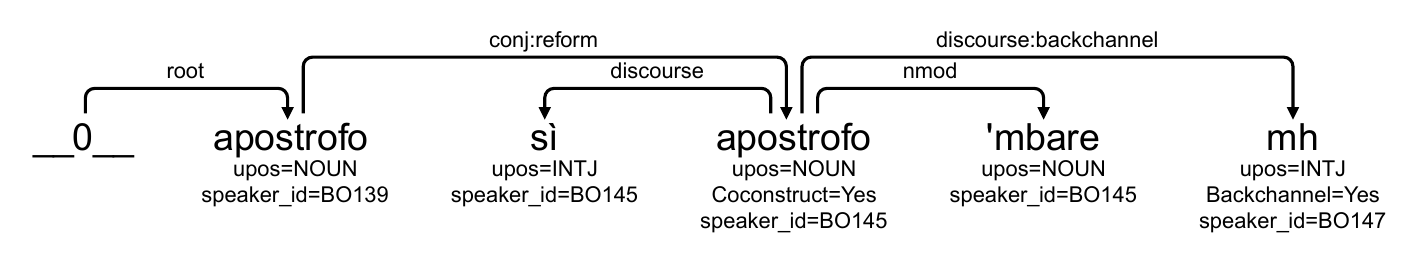}
  \caption{\label{fig:apostrofo}An Italian example of a \texttt{conj:reform} coconstruction and a backchannel. The example is taken from KIParla, BOA3017. Tr. en. `apostrophe', `yes, apostrophe \textit{'mbare}', `mh'. \textit{'mbare} is a dialectal form for \textit{compare}, `mate'.}
\end{figure*}

In the Figure~\ref{fig:apostrofo}, the first line is the speaker-based view and the last line is the dependency-based view.
An intermediate richer structure is introduced in which: new dependency relations are identified with a suffix \texttt{/attach} and they are drawn in green.
The richer structure helps to understand the conversion process and adds the possibility to query these modified dependencies specifically.
Note also that metadata relative to the speaker, which would usually be placed at sentence level, is now moved to tokens, to take into account the fact that a dependency-based sentence can include tokens uttered by different speakers.

\begin{figure*}
  \centering
  \includegraphics[scale=0.45]{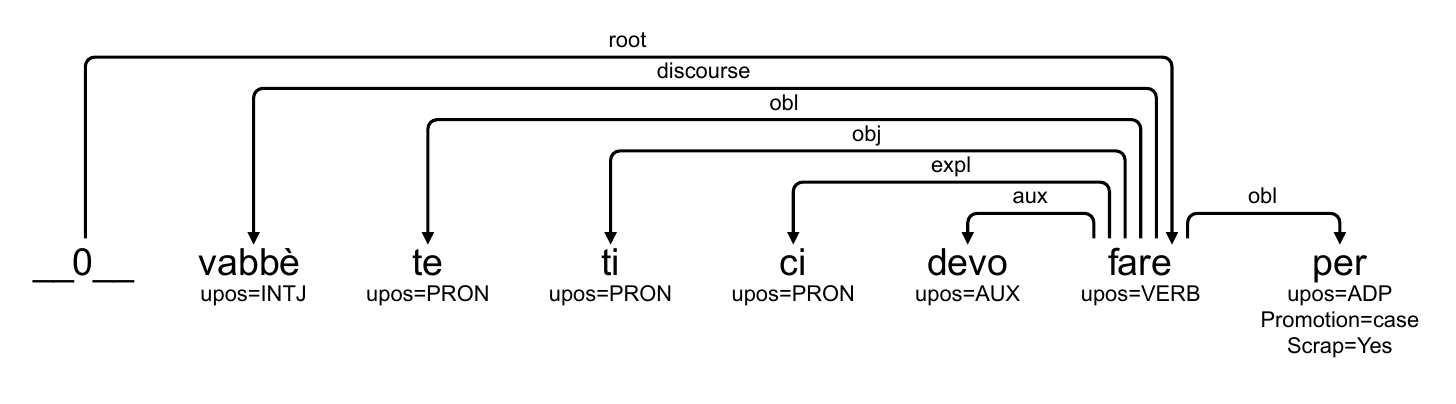}
  \includegraphics[scale=0.45]{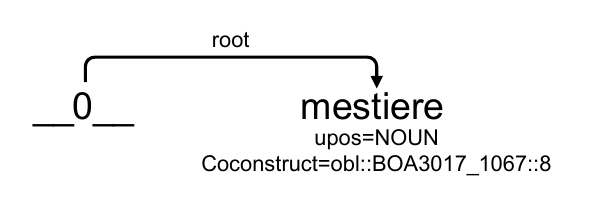}\\
  \includegraphics[scale=0.43]{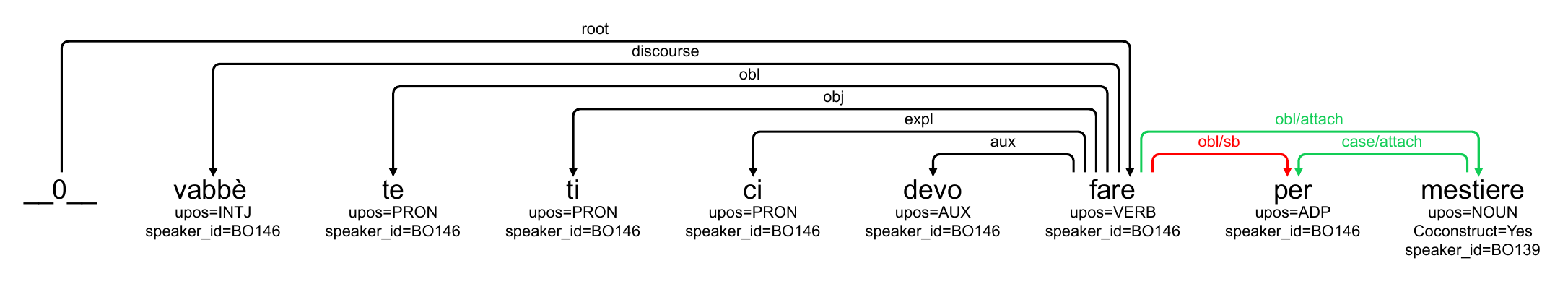}\\
  \includegraphics[scale=0.43]{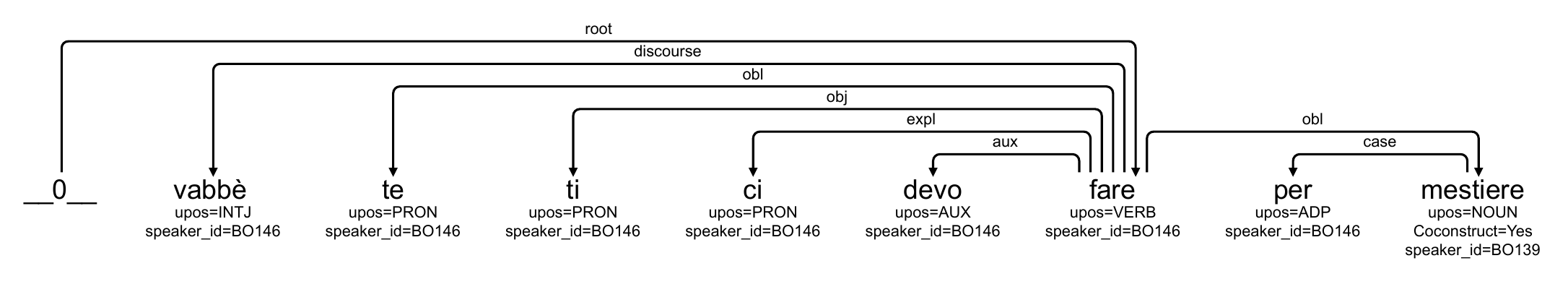}
  \caption{\label{fig:mestiere}Italian example~(\ref{ex:mestiere}) with promotion (NB: the sentence is shortened, for space reasons). The first line shows speaker-based annotation, the second line shows the intermediate graph representation of the dependency-based view and line 3 shows final UD well-formed tree for the dependency-based view.}
\end{figure*}

As explained above, there may be other dependency relations needing to be restored in dependency-based view.
The new features \texttt{Promotion} and \texttt{Head} are used to reconstruct the expected dependency structure.
The Figure~\ref{fig:mestiere} shows an Italian example where the oblique prepositional phrase \emph{per mestiere} is built by two different speakers.
In this case, we mark, on \emph{per} that the structure is unfinished (\texttt{Scrap=Yes}) and that it has been promoted (\texttt{Promotion=case}).
The second line in the figure show the intermediate (graph) structure where both dependencies on \emph{per} are drawn:
in red with suffix \texttt{sb} for the promoted relation and in green with suffix \texttt{attach}, the expected UD relation dependency-based view.
The last line gives the UD regular annotation of the dependency-based view.
Similar examples are given in appendix for English (Figure~\ref{fig:nice}) and for Slovenian (Figure~\ref{fig:sl}).

Note that when the promoted token has its own dependencies, they may have to be lifted up in the final dependency-based structure.
The figure~\ref{fig:nice} gives such an example on a built English example \emph{this is pretty} in which \emph{pretty} is a separate maximal unit in speaker-based view.

\section{Coconstructions in existing UD treebanks}\label{sec:treebanks}

\subsection{French (Rhapsodie)}
\label{sec:rhapsodie}

Coconstructions and backchannels were already annotated in the first version of Rhapsodie~\citep{benzitoun2010tu,lacheret2014rhapsodie}, using the delimiter //+, as in Example~(\ref{ex:bourgeoise}). When the treebank has been converted into UD, they were lost, and then reintroduced with a feature \texttt{AttachTo} for the address of the governor and \texttt{Rel} for the relation \citep{kahane2021annotation}. For backchannels, the value of \texttt{Rel} was \texttt{discourse}. Backchannels in Rhapsodie currently include \textit{yes-no} answers to questions.
Instead of \texttt{conj:reform} and \texttt{reparandum}, seven different labels were first considered \citep{kahane_typologie_2012}, but due to a quite poor inter-annotator agreement, they were merged into a unique relation, i.e., \texttt{conj:dicto}.

Among the 3,209 trees in the speaker-based version of Rhapsodie, 229 are annotated as backchannels, but there are also 93 interjections as roots that are potential backchannels and need to be revised. The treebanks also contain 35 coconstructions: 24 \texttt{conj:reform} and 11 modifiers, including two relative clauses.

\subsection{Italian (KIParlaForest)}
\label{sec:kiparla}

KIParla Forest~\citep{Pannitto_2025a} was first introduced in UD v2.17, based on the broader KIParla Corpus of Spoken Italian~\citep{Mauri_2019,BallareMauri2020}.

The treebank currently hosts 2 conversations from the original corpus, an interview (BOD2018, comprising 2 speakers) and a free conversation (BOA3017, comprising 4 speakers), for a total of $9,135$ tokens and $1,007$ sentences.

Backchannels are annotated in 134 sentences: most of these are sentences composed by one single token that is frequently an \texttt{INTJ} such as \textit{ah}, \textit{eh}, \textit{mh} or \textit{ok} (67 cases), an \texttt{ADV} as \textit{sì} `yes', \textit{no}, \textit{certo} `certainly', \textit{ecco} `that's it', \textit{esatto} `exactly' (51 cases). Another significant group is that of exact repetitions (4 cases)
Also exclamations such as \textit{porca troia} `holy shit' are treated as cases of backchannels, and longer sequences such as \textit{ah me l'hai raccontato} `right you told me'.

Currently the treebank contains 70 annotations of coconstructions.
Most cases (38) concern conjuncts, either proper \texttt{conj} or \texttt{conj:reform} relations. \texttt{obl} and \texttt{acl:*} also constitute a big group (13 cases), as they capture the prototypical case of expansion of the dependency tree during interaction.

\subsection{Slovenian (SST)}

In the original compilation of the Spoken Slovenian Treebank (SST, \citealt{dobrovoljc-nivre-2016-universal,Dobrovoljc2025-mn}), coconstructions were not treated explicitly. The treebank followed the speaker-based utterance segmentation of the GOS corpus~\citep{verdonik-etal-2024-gos}, which encompasses 6,121 utterances and 98,393 tokens, and syntactic annotation was applied independently within each utterance. To evaluate the feasibility of the proposed guidelines and estimate the extent of coconstructions in Slovenian spoken data, we therefore began by targeting two configurations: completion of an unrealized dependency (Type A.1 in Section~\ref{sec:coconstruction-proper}) and backchannelling (Section~\ref{sec:backchanneling}).

Backchannels were identified semi-automatically by extracting consecutive utterance pairs by different speakers where the first was not interrogative and the second contained one or more backchannel candidates (based on a preexisting list derived from items already marked as \texttt{discourse}, \texttt{INTJ} and/or \texttt{PART}). This procedure yielded 396 cases of backchanneling (69 different types), mostly single words (e.g. \textit{ja} 'yes', \textit{mhm} 'mm-hm', \textit{aha} 'I see', \textit{aja} 'oh, right', \textit{dobro} 'okay'), but also longer sequences (e.g. \textit{tako, tako, tako, tako} 'right, right, right, right').

For coconstruction proper (type A.1), we manually inspected utterance pairs where the first did not end in final punctuation, indicating potential incompletion, and manually identified 17 cases of completed projected dependencies, including argument insertion (Speaker 1: \textit{če atom odda elektrone, nastane...} `if an atom gives away electrons, there arises...' – Speaker 2: \textit{...kation} `...a cation') and sentence-level completion (Speaker 1: \textit{kdor ti bo reklamo poslal...} `whoever sends you an advertisement...' – Speaker 2: \textit{...zaračuna} `...charges money’).

These results provide an initial empirical basis for studying coconstructions in Slovenian and a retrieval procedure that can be refined to capture additional types, or extended to other UD treebanks where such phenomena have not yet been explicitly annotated.

\section{Conclusion}

UD has been first developed on the basis of written corpora with the aim of developing parsers~\citep{mcdonald2013universal}. The collection now includes spoken data for languages with a written tradition (as the three languages considered in this paper), as well as languages with only a spoken tradition (including sign languages). Due to the lack of clear guidelines for spoken data, important discrepancies exist among treebanks for spoken data~\citep{dobrovoljc_spoken_2022}. This work is part of a common effort to propose guidelines and unify the annotation of different treebanks.
The first step in the syntactic analysis of spoken data is segmentation into minimal units (words or morphemes) and maximal units. The latter question is the main topic of this paper and  the question of coconstructions is one of the very first problems to be solved. We hope to have convinced the reader of the interest in the question, as well as the relevancy of our propositions.

While the framework has been developed with spoken interaction in mind, its underlying mechanisms are more broadly applicable. Two extensions are worth noting. First, the proposed annotation scheme is potentially relevant for computer-mediated communication (CMC) data — such as instant messaging, or platforms like Slack or Discord — where users frequently fragment a single syntactic unit across multiple messages by sending partial utterances. These \emph{split utterances} are structurally analogous to unfinished phrases in speech, and the \texttt{Coconstruct}, \texttt{Scrap}, and \texttt{Promotion} features could be applied to capture cross-message syntactic dependencies. A distinctive challenge of CMC, however, is the presence of \emph{lag}: responses may not appear chronologically adjacent to the message they depend on, due to interleaved threads or parallel conversations. The pointer-based design of our features (encoding the identity of the eliciting token rather than relying on adjacency) makes the framework robust to such non-linear environments. Second, even within spoken data, the mechanism may be useful for cases of same-speaker continuation across deliberately maintained turn boundaries — for instance, when a speaker resumes a syntactic structure after a long backchannel sequence and the annotator wishes to preserve segmentation boundaries for prosodic or discourse reasons. In both cases, the features introduced here provide a lightweight but expressive mechanism for encoding syntactic continuity without collapsing the original segmentation.

We submit this proposal to the UD community hoping that it will be adopted as a standard extension for spoken and interactive treebanks, contributing to a more linguistically adequate treatment of multi-party and mediated language data.

\section*{Acknowledgements}
This work received support from the CA21167 COST action UniDive, funded by the European Union via the COST (European Cooperation in Science and Technology), and from the ARIS research programme P6-0411.

\section*{References}
\bibliographystyle{lrec2026-natbib}
\bibliography{coconstruct}

\appendix

\section{Annotation examples}
\label{sec:appendix}

\begin{figure}[h]
    \centering
    \includegraphics[width=\linewidth]{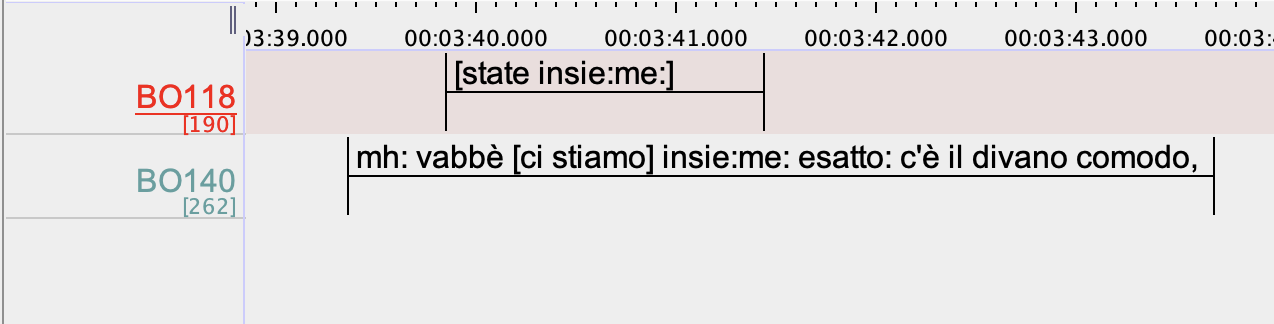}
    \caption{Original transcription of Example~(\ref{ex:mestiere}) in the KIParla corpus.}
    \label{fig:stiamo insieme}
\end{figure}

\begin{figure*}
  \centering
  \includegraphics[scale=0.5]{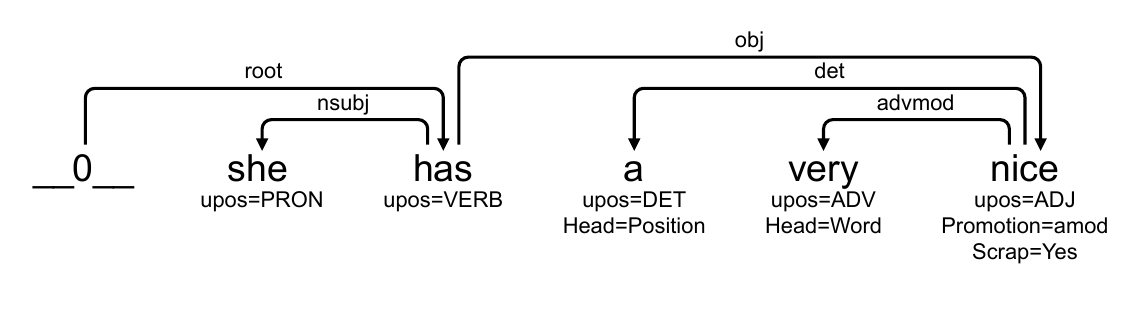} \qquad
  \includegraphics[scale=0.5]{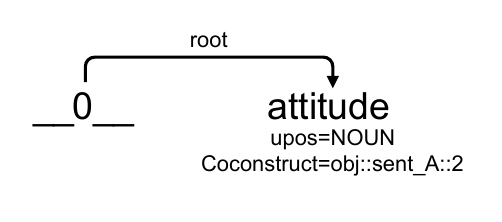}\\
  \includegraphics[scale=0.5]{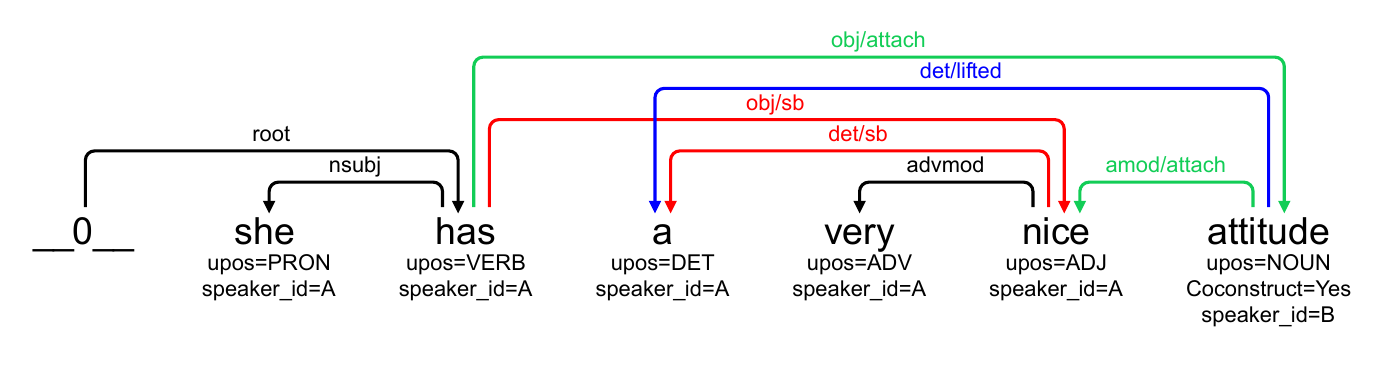}\\
  \includegraphics[scale=0.5]{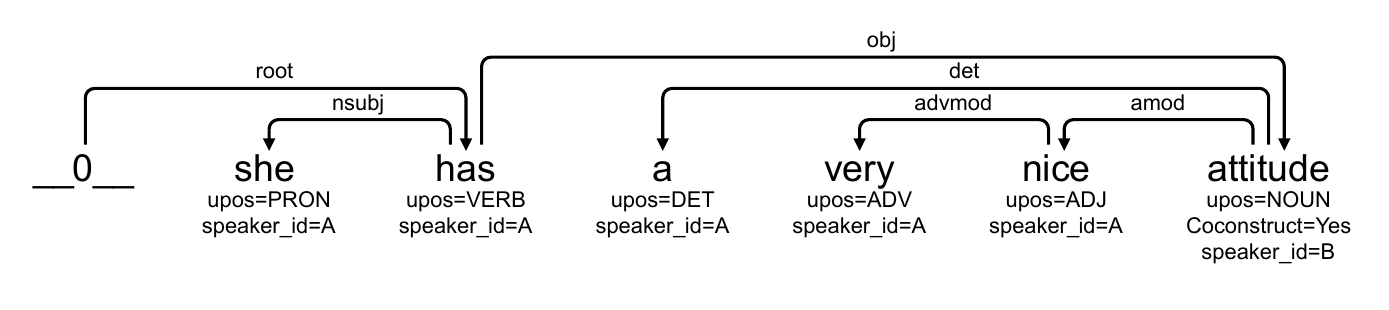}
  \caption{\label{fig:nice}English example with lifting.\\
  {Line 1: speaker-based annotation;}\\
  {Line 2: intermediate representation of the dependency-based view;}\\
  {Line 3: final UD tree for the dependency-based view}}
\end{figure*}

\begin{figure*}
  \centering
  \begin{minipage}{0.2\textwidth}
    \centering
    \includegraphics[angle=90,scale=0.45]{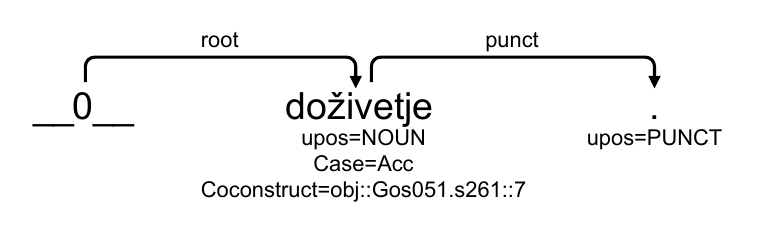}\\
    \includegraphics[angle=90,scale=0.45]{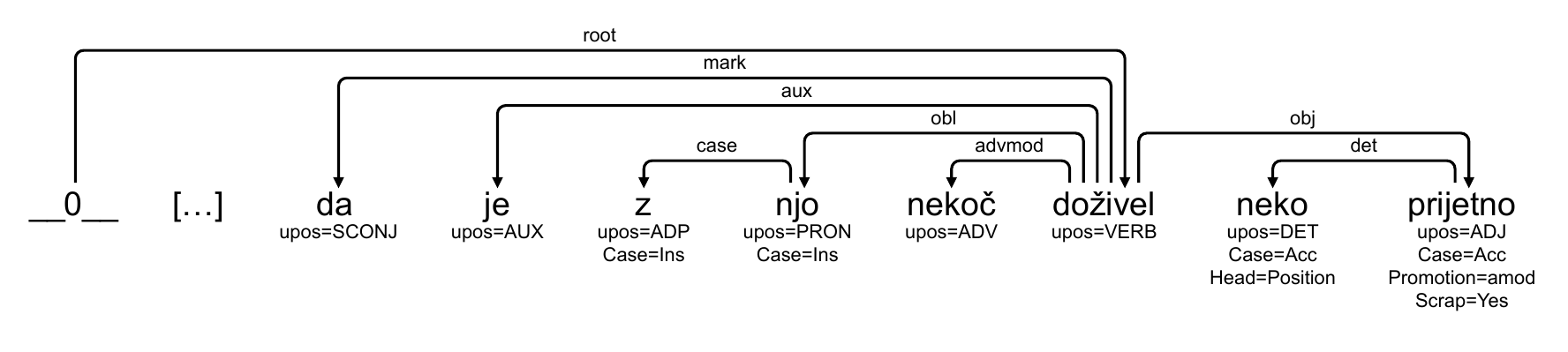}
  \end{minipage}
  \begin{minipage}{0.2\textwidth}
    \centering
    \includegraphics[angle=90,scale=0.45]{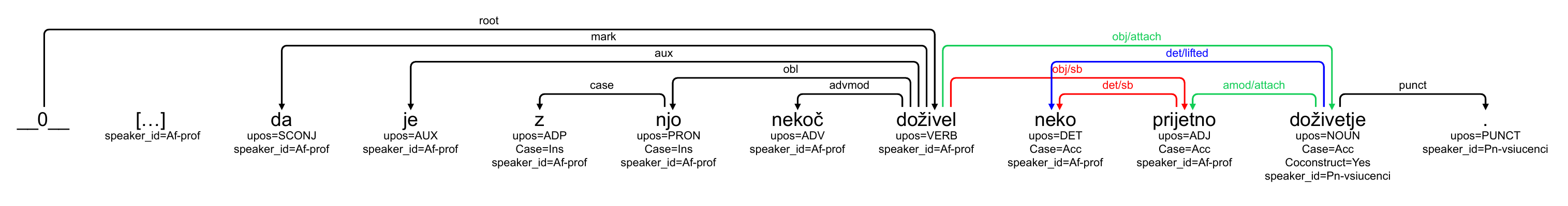}
  \end{minipage}
  \begin{minipage}{0.2\textwidth}
    \centering
    \includegraphics[angle=90,scale=0.45]{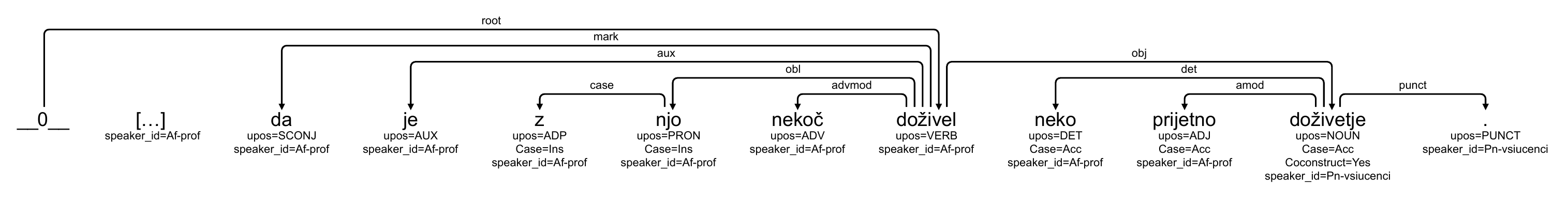}
  \end{minipage}
  \caption{\label{fig:sl}Slovenian example of promotion\\\footnotesize
  ‘[…] that he once had with her a pleasant … experience.’}
\end{figure*}

\begin{figure*}
  \centering
  \begin{minipage}{0.2\textwidth}
    \centering
    \includegraphics[angle=90,scale=0.45]{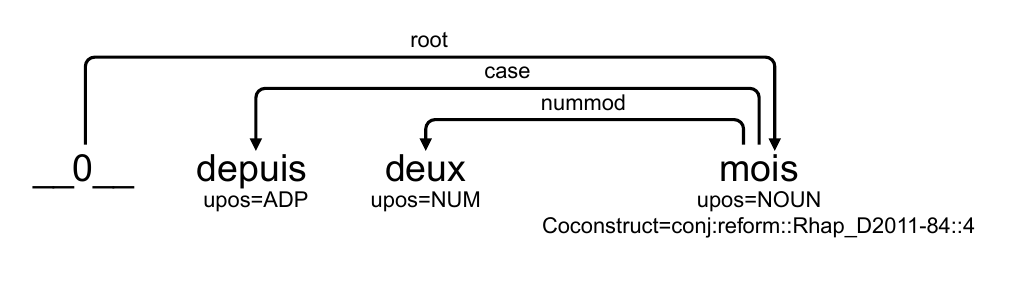}\\
    \includegraphics[angle=90,scale=0.45]{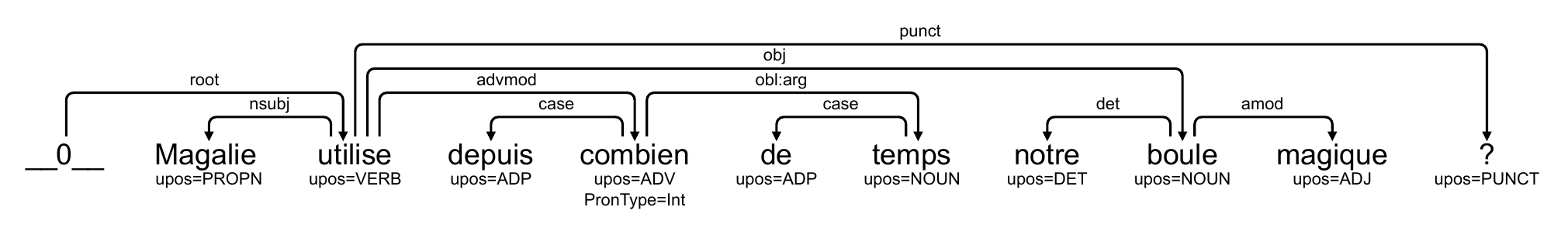}
  \end{minipage}
  \begin{minipage}{0.2\textwidth}
    \centering
    \includegraphics[angle=90,scale=0.45]{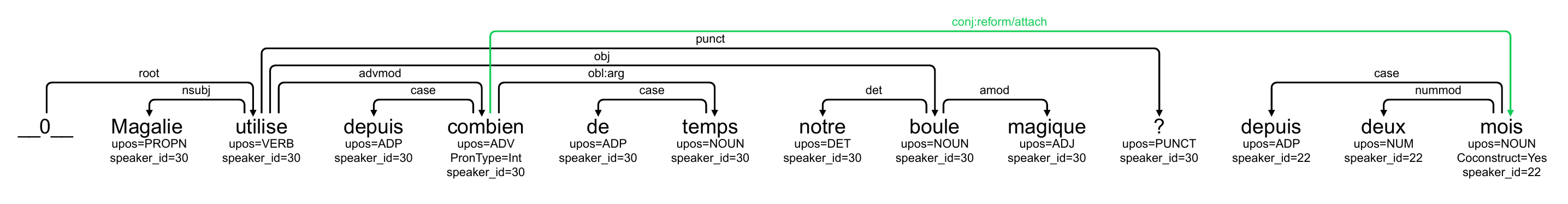}
  \end{minipage}
  \begin{minipage}{0.2\textwidth}
    \centering
    \includegraphics[angle=90,scale=0.45]{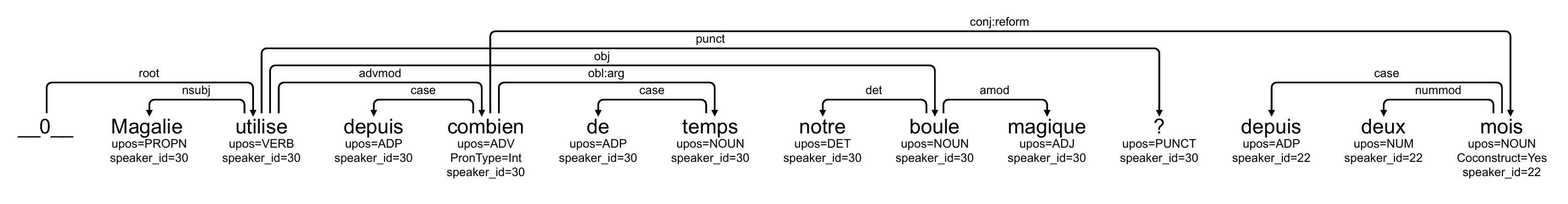}
  \end{minipage}
  \caption{\label{fig:boule}French example with a question\\\footnotesize
  ‘How long has Magalie been using our magic ball?', `for two months’}
\end{figure*}

\begin{figure*}
  \centering
  \begin{minipage}{0.2\textwidth}
    \centering
    \includegraphics[angle=90,scale=0.45]{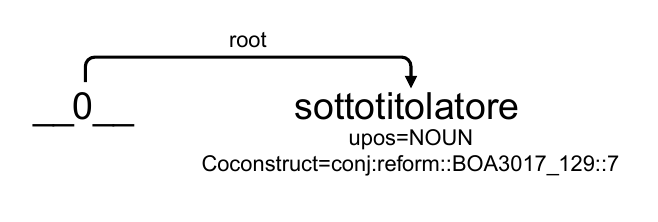}\\
    \includegraphics[angle=90,scale=0.45]{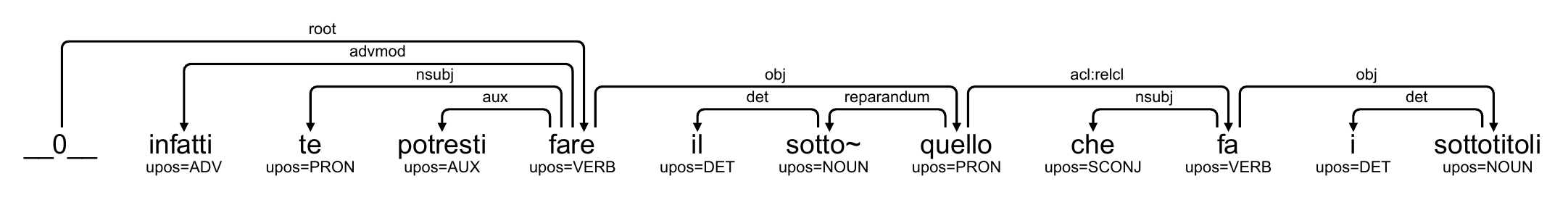}
  \end{minipage}
  \begin{minipage}{0.2\textwidth}
    \centering
    \includegraphics[angle=90,scale=0.45]{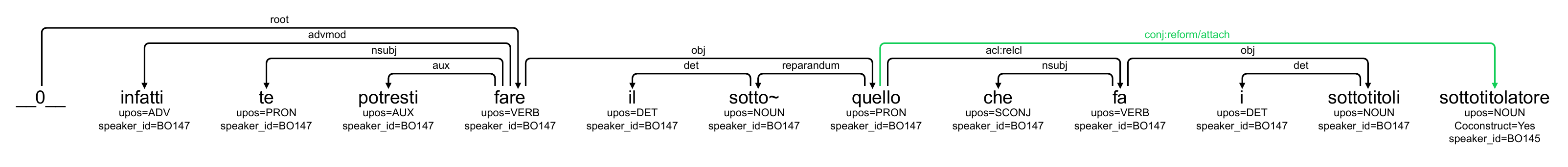}
  \end{minipage}
  \begin{minipage}{0.2\textwidth}
    \centering
    \includegraphics[angle=90,scale=0.45]{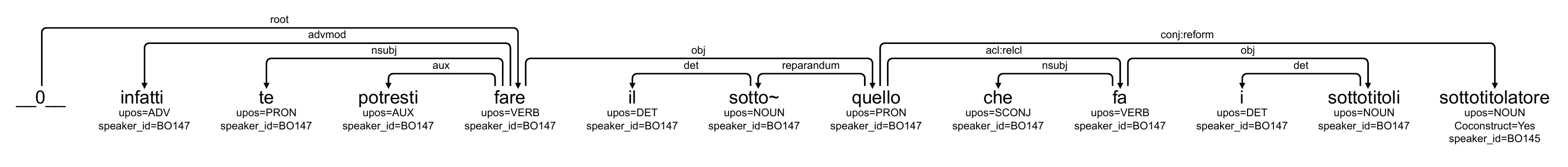}
  \end{minipage}
  \caption{\label{fig:repair}Italian repair example, see Example~(\ref{ex:sottotitolo}) for translation.}
\end{figure*}

\end{document}